\documentclass[11pt,a4paper,twocolumn,logo]{evomem3dgoogle}
\usepackage{multirow}
\usepackage{subcaption}
\usepackage{siunitx}
\usepackage{array}
\usepackage[authoryear,round]{natbib}
\let\cite\citep
\usepackage[capitalize,noabbrev,nameinlink]{cleveref}
\usepackage{catchfile}
\usepackage{caption}

\ifXeTeX\microtypesetup{tracking=false}\fi

\newcommand{\method}{\textsc{EvoMem3D}}
\newcommand{\early}{\textsc{Early}}
\newcommand{\intermediate}{\textsc{Intermediate}}
\newcommand{\lateexit}{\textsc{Late}}
\newcommand{\fullexit}{\textsc{Full}}

\newcommand{\cmark}{\ding{51}}
\newcommand{\xmark}{\ding{55}}

\title{EvoMem3D: Adaptive-Depth Temporal Representation Memory for Multi-Camera 3D Detection}
\author{Moshe Kimhi}
\author{Idan Kligvasser}
\author{Yotam Intrator}
\author{Regev Cohen}
\author{Ehud Rivlin}
\affil{Google Research}
\correspondingauthor{\texttt{moshekimhi@google.com}}
\makeatletter
\g@addto@macro\ps@firststyle{%
  \fancyfoot[L]{\footerfont\textit{Corresponding author: }
    \texttt{moshekimhi@google.com}}}
\makeatother

\renewcommand{\today}{}
\renewcommand{\monthyeardate}{}
\hypersetup{pdftitle={EvoMem3D: Adaptive-Depth Temporal Representation Memory for Multi-Camera 3D Detection},
  pdfauthor={Moshe Kimhi, Idan Kligvasser, Yotam Intrator, Regev Cohen, Ehud Rivlin}}

\long\def\EvoSplitAbstract#1\begin#2#3\end#4#5\EvoAbstractEnd{%
  \gdef\theabstract{#3}%
  \gdef\EvoAfterAbstract{#5}}
\expandafter\EvoSplitAbstract

\begin{abstract}
Modern multi-camera 3D detectors increasingly rely on large
Vision Transformer (ViT) image encoders, making repeated
full-depth encoding a dominant inference cost even though
adjacent driving frames are highly redundant. We present
\method, an adaptive-depth temporal framework that caches
the most recent full-depth detector-interface appearance
representation as a read-only anchor and combines it with
policy-selected partial features from the current frame.
A lightweight pre-encoding controller chooses among semantic
encoder exits before image processing, allocating shallow
computation to stable frames and deeper computation when the
visual state changes. To reuse appearance without carrying
stale geometry, we decouple temporal appearance from geometric
alignment: current-frame calibration defines the decoder
geometry at every timestamp, while a motion-conditioned
positional adapter accounts for anchor-to-current ego motion.
On nuScenes, \method{} achieves 61.50 NDS and 52.76 mAP while
reducing end-to-end latency from 364.4 to 168.7 ms
(2.16$\times$) relative to standard dense DETR3D. With
StreamPETR, it reaches 61.36 NDS and 52.55 mAP at a
1.92$\times$ speedup. Controlled studies isolate the effects
of current evidence, temporal reuse, memory correction, and
adaptive routing, while our audit confirms that the policy learns to distinguish semantic driving regimes.
\end{abstract}

\begin{figure}
    \centering
    \includegraphics[width=\linewidth]{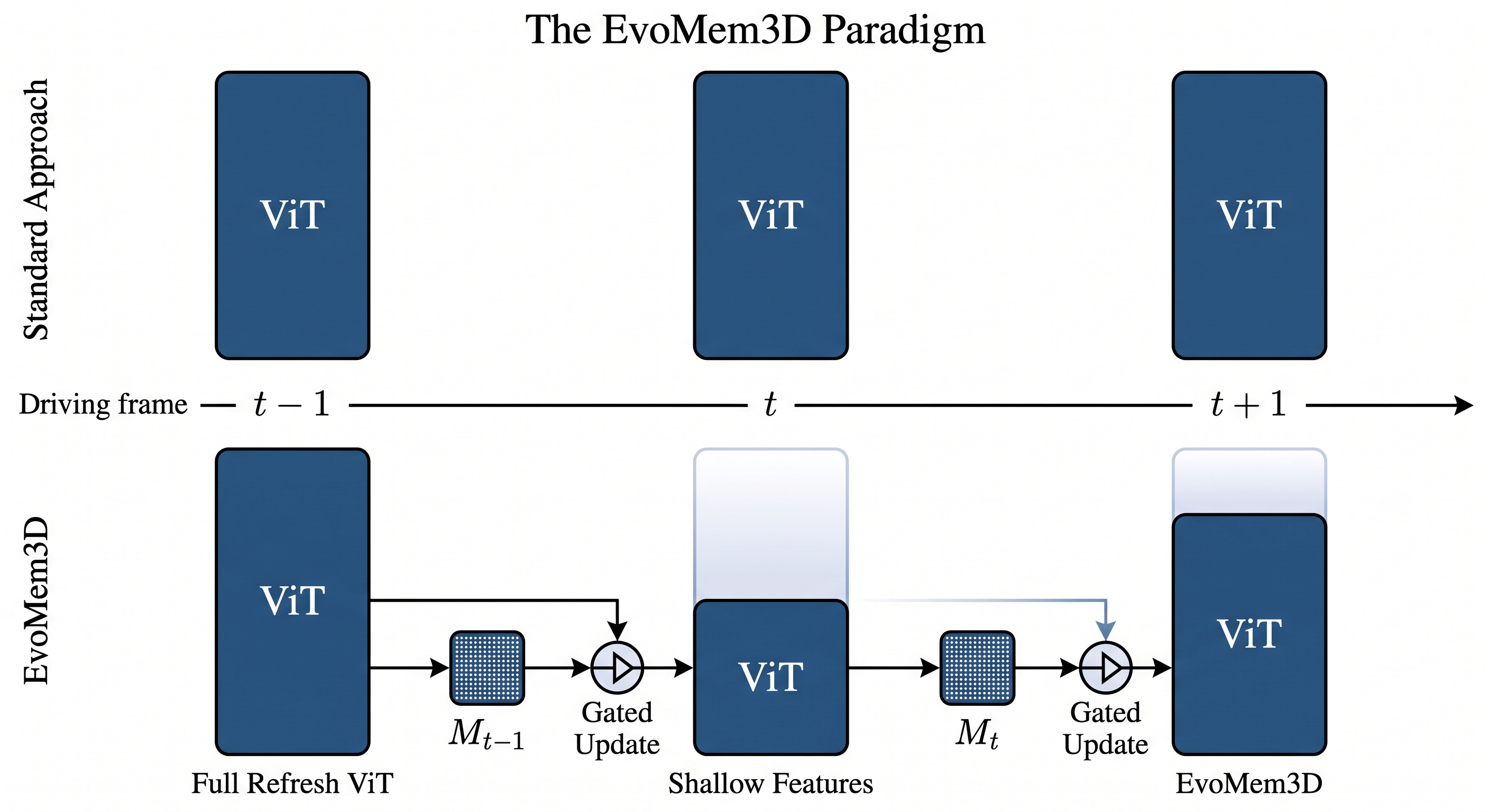}
    \caption{\textbf{Adaptive-Depth Temporal Representation Memory.} While standard multi-camera 3D detectors execute a full-depth vision transformer for every frame, \method{} treats the dense detector-interface representation as a recurrent memory. A lightweight policy dynamically allocates transformer depth based on ego motion and scene change, updating the persistent high-level memory with shallow current-frame evidence.}
    \label{fig:teaser}
\end{figure}

\EvoAbstractEnd

\let\EvoOriginalAbscontent\abscontent
\renewcommand{\abscontent}{%
  \EvoOriginalAbscontent
  \par\medskip\noindent
  {\small\textbf{Project page:} \url{https://mkimhi.github.io/EvoMem3D/}}}

\begin{document}
\maketitle
\EvoAfterAbstract

\begin{figure*}
    \centering
    \includegraphics[width=\linewidth, trim=15 102 5 53, clip]{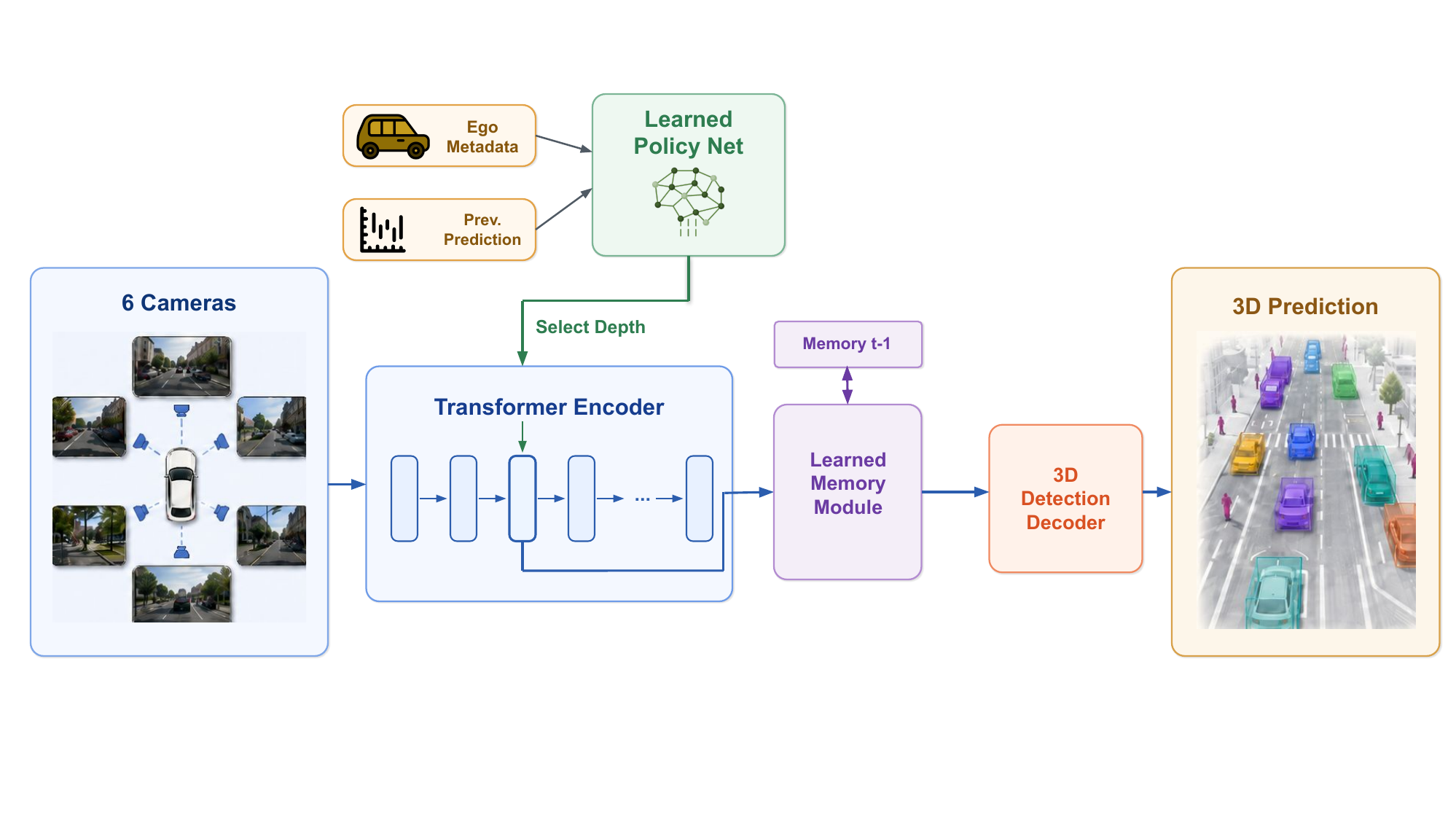}
\caption{\textbf{Method overview.}
Before image encoding, the controller selects a semantic exit using
deployable temporal signals. A full-depth appearance representation
serves as the temporal reference; partial actions combine it with
projected current features to produce $\widetilde{A}_t$. Relative ego
motion conditions only the current decoder input.}
    \label{fig:overview}
\end{figure*}

\section{Introduction}
\label{sec:intro}

3D object detection is a central perception problem for
autonomous driving. LiDAR-based detectors exploit explicit
3D geometry through voxel, pillar, and center-based
representations~\cite{zhou2018voxelnet,lang2019pointpillars,
yin2021centerpoint}, while camera-based methods infer 3D
structure from visual observations using monocular,
perspective-view, or bird's-eye-view representations
~\cite{wang2021fcos3d,philion2020liftsplatshootencoding,li2023bevdepth}.
Multi-camera detectors increasingly rely on large image
encoders to produce strong visual features. Detector-side
temporal mechanisms can propagate object queries or BEV state,
but they usually do not reduce the cost of the image backbone:
every timestamp is still encoded at full depth. This is
inefficient in driving video, where adjacent observations
contain substantial redundant visual information.

A direct early exit is not enough. An early transformer representation contains current visual evidence, but it does not have the semantic depth expected by a detector trained on final features. Conversely, reusing a previous final representation preserves semantic content but becomes stale as the cameras and scene move. The useful operating point lies between these extremes: retain a full high-level representation as temporal context, and use partial computation on the current frame to correct it.

\cref{fig:teaser} illustrates this operating point: the most recent
full-depth representation remains a stable appearance anchor,
while current frames spend only the encoder depth needed to
supply fresh visual evidence.

We introduce \method{}, an adaptive temporal feature-memory framework for multi-camera 3D detection. The method intercepts the dense appearance representation between the image encoder and the geometry-aware detection head. A full action replaces the cached representation. A partial action projects the selected intermediate feature to the same interface and fuses it with the cached full representation. The resulting feature is consumed by the unchanged detector head using current calibration and ego pose. This separation avoids propagating stale positional encoding through memory.

The amount of partial computation should depend on the frame. Stable intervals may require only early visual evidence, while viewpoint changes or new objects may require deeper processing or a full refresh. \method{} therefore uses a pre-encoding controller that selects from \{\early{}, \intermediate{}, \lateexit{}, \fullexit{}\}. The controller is trained with an ordered representation-change target derived from full features during training. At inference, it uses only information available before processing the current image, including ego-motion history, memory state, and summaries of the preceding detector output.


Our contributions are:
\begin{itemize}[leftmargin=*]
    \item We formulate dense pre-geometry image features as temporal memory that can be corrected with partial current-frame computation without modifying the detector head.
    \item We introduce a pre-encoding controller over semantic depth levels, trained from full-representation change while using deployable temporal signals at inference.
    \item We decouple temporal appearance reuse from geometric alignment
through a motion-conditioned PE adapter: relative ego motion produces a
zero-initialized positional bias applied only to the current decoder
input, while the cached appearance anchor remains unchanged.
    \item We provide controlled accuracy, measured latency, and operation-count comparisons that isolate feature reuse, partial encoding, memory correction, update design, and adaptive routing, alongside an interpretability audit showing that the policy clusters clips into distinct driving behaviors.
\end{itemize}

\section{Related Work}
\label{sec:related}

\paragraph{Temporal camera-only 3D detection.}
Camera-only 3D detection spans monocular, perspective-view,
and BEV formulations. Lift-Splat-Shoot~\cite{philion2020liftsplatshootencoding}
established an influential explicit multi-camera lifting
formulation, while BEVDepth~\cite{li2023bevdepth} improves the
geometric reliability of BEV construction through explicit
depth supervision.
DETR3D samples multi-view image features through projected 3D reference points, and PETR encodes 3D coordinates directly in the perspective-view feature domain \cite{detr3d,liu2022petr}. Temporal variants aggregate history in bird's-eye view, as in BEVFormer and BEVDet4D \cite{li2022bevformer,huang2022bevdet4d}, or propagate object-query memory online, as in StreamPETR and SparseBEV \cite{wang2023streampetr,liu2023sparsebev}. In all of these detectors, temporal state lives downstream of the image backbone, so every frame is still encoded at full depth. \method{} moves temporal state into the encoder: the dense pre-geometry representation itself persists across frames and is corrected rather than recomputed.

\paragraph{Temporal feature reuse and adaptive video inference.}
Video models have long reduced repeated computation by evaluating
expensive features sparsely and propagating or updating them between
keyframes. Deep Feature Flow warps keyframe features using optical flow
\cite{zhu2017dff}, while Zhu et al.\ combine sparse keyframes with
partial feature updating and adaptive scheduling
\cite{Zhu_2018_CVPR}. FrameExit instead learns conditional early
termination for video recognition \cite{ghodrati2021frameexit}.
\method{} shares the objective of allocating computation over time,
but operates on synchronized multi-camera features consumed by a
geometry-aware 3D detector. Rather than warping a 2D feature or
recursively updating temporal state, it combines a read-only
full-depth appearance anchor with partial current-frame evidence.

\paragraph{Position encoding and temporal geometry.}
PETR maps 3D coordinates into perspective-view position embeddings
\cite{liu2022petr}, and PETRv2 extends this representation to align
object positions across frames \cite{Liu_2023_ICCV}. CAPE constructs
camera-view position embeddings and encodes ego motion for temporal
query modeling \cite{xiong2023cape}. Our contribution is orthogonal:
we do not transform the cached dense appearance representation.
Instead, anchor-to-current ego motion produces a zero-initialized
adaptation applied only to the current decoder positional input.

\paragraph{Efficient transformers for 3D detection.}
DynamicViT prunes tokens, AdaViT skips blocks, and A-ViT halts individual tokens adaptively per input \cite{rao2021dynamicvit,meng2022adavit,yin2022avit}. CARES reduce the token count by selecting smaller resolution \cite{cares,awares}. In multi-camera detection, ToC3D routes image tokens using historical object queries \cite{zhang2024toc3d}, SEPatch3D adapts image patch sizes to scene content \cite{SEPatch3D}, and SToRe3D jointly sparsifies image tokens and 3D queries \cite{store3d}. These methods decide how much to compute within a timestamp: the redundancy they exploit is spatial. \method{} decides across timestamps: full depth when the scene changes, a shallow prefix when it does not. The two families therefore answer different questions, and \cref{sec:experiments} compares \method{} against representative spatial methods in their own test environments.


\section{Method}
\label{sec:method}

\method{} augments a pretrained multi-camera 3D detector with a
temporal feature memory and a lightweight depth controller, leaving the detector itself unchanged. Full-depth features define the memory anchor, partial encodings correct it with current evidence, and the controller chooses the depth of each frame before its images are processed. We first give an overview of the framework, then describe the semantic exits and their projection (\cref{sec:method-exits}), temporal memory
correction (\cref{sec:memory}), motion-conditioned positional adaptation
(\cref{sec:motion_pe}), the pre-encoding controller (\cref{sec:method-controller}), and its
supervision (\cref{sec:method-supervision}).

\subsection{Overview of Adaptive-Depth Memory}
Standard multi-camera 3D detectors process synchronized camera images $I_t=\{I_t^v\}_{v=1}^{V}$ at timestamp $t$ through a static pipeline: an image backbone $F$, a feature pyramid neck $\Pi$, and a geometry-conditioned detection head $D$. \method{} replaces this static extraction with a dynamic temporal memory framework.

Instead of executing the full backbone for every frame, a lightweight controller first evaluates the temporal context and the ego car kinematics to select the encoding depth.
We maintain a dense, detector-interface representation cache, $C_t$, holding the visual features from the most recent full encoding. When a partial action is selected, the model extracts intermediate backbone features, projects them into the same dimensional space as the cache, and fuses them to correct stale visual evidence. The geometry-aware 3D head remains completely unchanged, consuming this updated representation alongside the current frame's precise positional and camera calibration data ($\gamma_t$).
The complete data flow is summarized in \cref{fig:overview}.

\subsection{Adaptive Exits and Feature Projection}
\label{sec:method-exits}
To allow intermediate representations to interact seamlessly with the full-depth cache, we define a unified memory space. A standard full-depth computation yields a multi-scale appearance representation:
\begin{equation}\label{eq:A}
    A_t^\star = \Pi_{\mathrm{full}}(F_{\mathrm{full}}(I_t)),
\end{equation}
where $F_{\mathrm{full}}$ is the complete image encoder and $\Pi_{\mathrm{full}}$ is the standard FPN neck \cite{fpn}.

For a partial encoded action $a_t$, we halt the backbone early to extract intermediate features $F_{a_t}(I_t)$. Because these intermediate features reside in different semantic and dimensional spaces, each exit is equipped with a dedicated, lightweight projector (a specialized FPN), $\Pi_{a_t}$. This projects the partial features directly into the canonical detector-interface format:
\begin{equation}
    P_t^{a_t} = \Pi_{a_t}(F_{a_t}(I_t)).
\end{equation}

\subsection{Temporal Memory and Correction}
\label{sec:memory}

Let $C_t$ denote the recurrent appearance memory available
before frame $t$, and let $\tau_t$ denote the timestamp of
its most recent full-depth refresh. A full-depth action
produces the representation $A_t^\star$ defined in \cref{eq:A}
and refreshes the memory directly.

For a partial action $a_t$, the projected current-frame
feature $P_t^{a_t}$ is combined with the current memory
$C_t$ by a lightweight update operator $U$. At
feature-pyramid level $\ell$, we instantiate this operator as
a gated residual:

\begin{equation}
\widetilde{A}_t^{(\ell)}
=
C_t^{(\ell)}
+
\alpha_\ell
g_\ell\!\left(
    [C_t^{(\ell)};P_t^{a_t,(\ell)}]
\right)
\odot
r_\ell\!\left(
    [C_t^{(\ell)};P_t^{a_t,(\ell)}]
\right),
\end{equation}

where $g\ell$ is sigmoid-bounded, $r_\ell$ predicts a
residual, and $\alpha_\ell$ is a learned residual scale.
Collecting the pyramid levels defines
$U(C_t,P_t^{a_t})$. The updater is shared across partial
exits, while each exit has its own projector into this common
feature space.

The representation used by the detector and the recurrent
memory therefore evolve jointly as
\begin{equation}
(\widetilde{A}_t,C_{t+1},\tau_{t+1})
=
\begin{cases}
(A_t^\star,A_t^\star,t),
    & a_t=\textsc{Full},\\[2mm]
(U(C_t,P_t^{a_t}),C_t,\tau_t),
    & \text{otherwise}.
\end{cases}
\end{equation}
Thus, on a partial frame, shallow current evidence both
corrects the representation used for the current prediction
and updates the memory passed to the next timestamp.
Successive partial actions therefore refine the recurrent
memory over time. A \textsc{Full} action refreshes this state
with a full-depth current representation; $\tau_t$ tracks
the most recent such refresh and is used to describe the
age and accumulated motion of the full-depth reference.

\begin{figure*}
    \centering
    \includegraphics[width=0.9\textwidth]{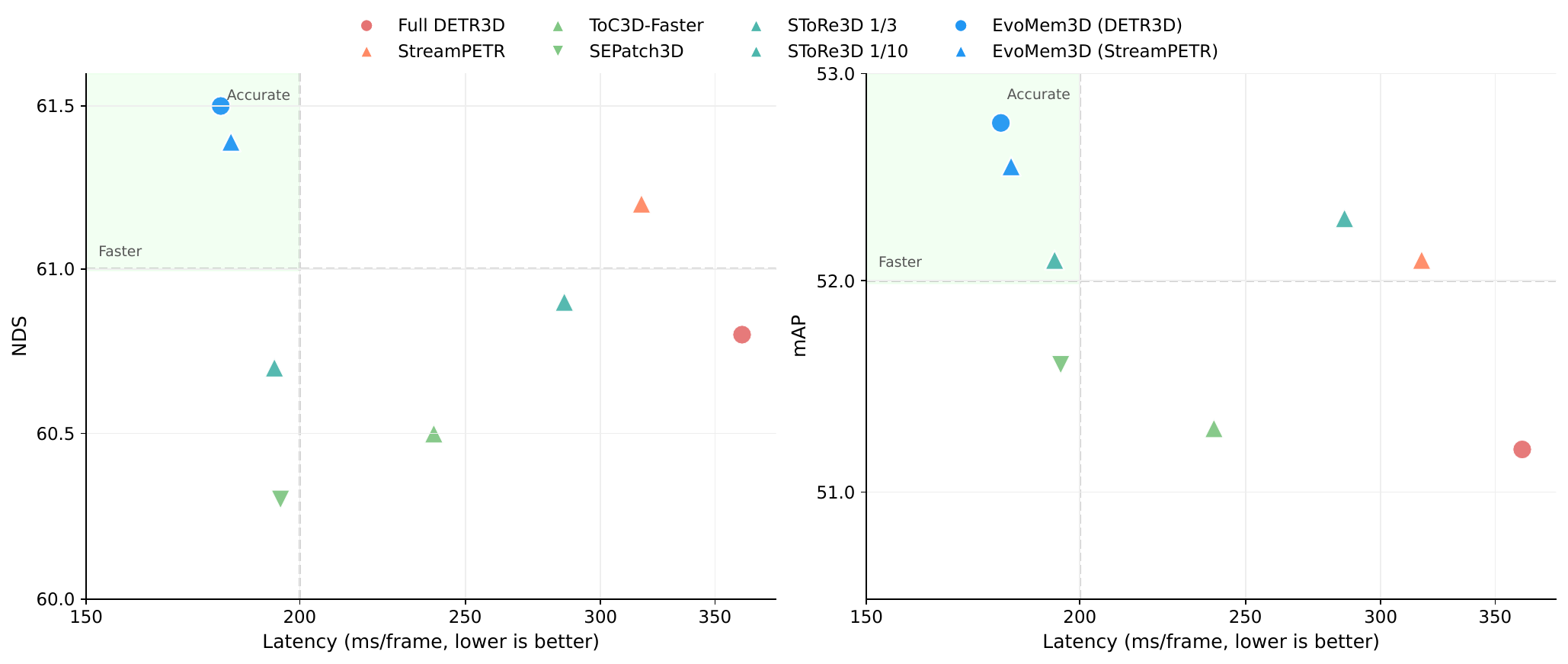}
    \caption{Accuracy--latency trade-off comparing methods using ViT-L. Triangle is based on StreamPETR, and circle based on DETR3D base model.}
    \label{fig:accuracy-latency}
\end{figure*}

\subsection{Motion-Conditioned Positional Adaptation}
\label{sec:motion_pe}
When appearance is reused across timestamps, its reference pose may
differ from the current camera pose. We keep appearance reuse and
geometric alignment separate: the cached tensor is treated purely as
visual context, while temporal geometry is expressed through a
lightweight correction to the positional encoding consumed by the
geometry-aware decoder.

Using the anchor timestamp $\tau_t$ defined above, let $T_s$ denote
the global ego pose at time $s$. The relative transform from the
anchor ego frame to the current ego frame is
\begin{equation}
\Delta T_t = T_t^{-1}T_{\tau_t}.
\end{equation}
We summarize the accumulated motion by
\begin{equation}
\psi(\Delta T_t,\Delta t_t)
=
[\Delta\mathbf{x}_t,
 \sin(\Delta\theta_t),
 \cos(\Delta\theta_t),
 \Delta t_t,
 v_{xy},
 \omega_z],
\end{equation}
where $\Delta t_t=t-\tau_t$, $\Delta\theta_t$ is the relative yaw,
and
\begin{equation}
v_{xy}
=
\frac{\sqrt{\Delta x_t^2+\Delta y_t^2}}
     {\max(\Delta t_t,\epsilon)},
\qquad
\omega_z
=
\frac{\Delta\theta_t}
     {\max(\Delta t_t,\epsilon)}.
\end{equation}
sine--cosine rep. avoids the yaw discontinuity at
$\pm\pi$.

A lightweight motion encoder maps this descriptor to
\begin{equation}
e_t=f_{\mathrm{mot}}\!\left(\psi(\Delta T_t,\Delta t_t)\right),
\end{equation}
which is projected to a positional correction
\begin{equation}
\delta p_t=\alpha W_{\mathrm{pe}}e_t.
\end{equation}
The correction is initialized near zero, allowing the
 detector to initially recover its original positional
encoding.

Let $p_t$ denote the detector's standard positional encoding,
constructed from the current camera calibration and geometry. When
temporal appearance is reused, the decoder receives
\begin{equation}
\widetilde{p}_t=p_t+\delta p_t,
\end{equation}
and predicts
\begin{equation}
\widehat{Y}_t
=
D\!\left(\widetilde{A}_t;
         \widetilde{p}_t,\gamma_t\right),
\end{equation}
where $\gamma_t$ denotes the remaining current-frame camera and
geometry metadata.

The motion correction is therefore frame-local: it modifies only the decoder positional input and is never written into the appearance
memory. When no temporal alignment is required, we simply use
$\delta p_t=0$.
\subsection{Pre-Encoding Depth Controller}
\label{sec:method-controller}
The controller evaluates the frame \textit{before} any current image encoding occurs. Its input is defined as:
\begin{equation}
    q_t = [m_t, \; h_t, \; \phi(\widehat{Y}_{t-1})],
\end{equation}
where $m_t$ contains timestamp and ego-motion history, $h_t$ describes the sequence state, and $\phi(\widehat{Y}_{t-1})$ summarizes the preceding detector output available at inference. A lightweight policy $\pi_\theta(a_t \mid q_t)$ predicts the depth action.

The controller also enforces validity constraints, overriding the policy to select \fullexit{} at scene starts or when the anchor age exceeds a strict bound (tightened further under high ego-motion). This ensures the temporal memory never becomes arbitrarily stale.

\subsection{Representation-Change Supervision}
\label{sec:method-supervision}
We supervise the policy using an ordered proxy for how much the full representation changes between adjacent frames. For scale $s$, camera $v$, and spatial location $u$, let $A_t^{\star,s,v}(u)$ be the channel vector of the full feature. The inter-frame similarity is:
\begin{equation}
    s_t = \frac{1}{|\mathcal{S}|} \sum_{s\in\mathcal{S}} \frac{1}{V|\Omega_s|} \sum_{v=1}^{V} \sum_{u\in\Omega_s} \cos\left(A_t^{\star,s,v}(u), A_{t-1}^{\star,s,v}(u)\right).
\end{equation}
Lower similarity indicates larger representation change. Ordered thresholds convert $s_t$ into semantic exit labels, assigning deeper computation to frames with higher visual drift. This target is computed only during training.

Policy training uses teacher forcing for the object-summary coordinates of $q_t$: $\phi(\widehat{Y}_{t-1})$ is replaced by the corresponding summary of the preceding ground-truth objects. At inference, the policy receives the preceding detector prediction. To reduce this train--inference shift, we standardize each policy coordinate using training-set statistics,

\begin{equation}
    x_t=\frac{q_t-\mu_q}{\sigma_q}
\end{equation}
where $\mu_q$ and $\sigma_q$ are coordinate-wise train set means and
s.t.d, with standard deviations clipped below by $\epsilon$.

and perturb the standarded policy input during training only:
\begin{equation}
    \widetilde{x}_t = x_t + \sigma_{\mathrm{n}}\epsilon_t,
    \qquad \epsilon_t \sim \mathcal{N}(0,I).
    \label{eq:policy-feature-noise}
\end{equation}
The reported model uses $\sigma_{\mathrm{n}}=0.10$ and evaluates the clean input $x_t$ at inference. This is policy-feature noise; the similarity target $s_t$ is not perturbed in the reported model.

Let $y_t$ denote the threshold-derived depth label and $w_{y_t}$ its class weight. The policy loss is the class-weighted multiclass cross-entropy averaged over valid supervised frames $\mathcal{T}_{\mathrm{sup}}$:
\begin{equation}
    \mathcal{L}_{\mathrm{pol}}
    = -\frac{1}{|\mathcal{T}_{\mathrm{sup}}|}
    \sum_{t\in\mathcal{T}_{\mathrm{sup}}}
    w_{y_t}\log \pi_\theta(y_t\mid\widetilde{x}_t).
    \label{eq:policy-loss}
\end{equation}
The labels are ordered by encoder depth, but the current implementation uses ordinary multiclass cross-entropy and does not impose an additional ordinal-distance penalty. At optimization step $k$, the objective is
\begin{equation}
    \mathcal{L}(k) = \mathcal{L}_{\mathrm{det}}
    + \lambda_{\mathrm{pol}}\,\mathbb{1}[k < K_{\mathrm{pol}}]\,
    \mathcal{L}_{\mathrm{pol}},
    \label{eq:training-objective}
\end{equation}
where $K_{\mathrm{pol}}$ is 10\% of the training schedule and $\lambda_{\mathrm{pol}}=1$ in the reported configuration. During this initial phase, the full encoder is evaluated without gradients on adjacent frames to construct $s_t$ and $y_t$; this teacher path is disabled afterward and is never used at inference. Because action selection is discrete, $\mathcal{L}_{\mathrm{det}}$ trains the partial projectors and memory updater but does not backpropagate through the selected action; the policy parameters are trained directly by $\mathcal{L}_{\mathrm{pol}}$.
After step $K_{\mathrm{pol}}$, the policy parameters are frozen and
only the partial projectors, updater, and positional adapter continue
to be optimized.

\setcounter{topnumber}{3}
\setcounter{totalnumber}{5}
\renewcommand{\topfraction}{0.9}
\renewcommand{\textfraction}{0.08}
\renewcommand{\floatpagefraction}{0.85}
\setlength{\floatsep}{10pt plus 2pt minus 2pt}
\setlength{\textfloatsep}{14pt plus 2pt minus 2pt}

\section{Experiments}
\label{sec:experiments}

We evaluate \method{} along five axes: (i) the accuracy--efficiency trade-off against prior efficient multi-camera detectors, (ii) the contributions of current evidence, temporal anchoring, and adaptive depth, (iii) whether full-representation change provides a useful supervisory signal for compute demand, (iv) the memory and latency overhead of the proposed mechanism, and (v) an interpretability audit of what driving regimes the pre-encoding policy learns to distinguish.

\subsection{Experimental Setup}

\paragraph{dataset and Metrics}
We evaluate our method on the nuScenes \cite{caesar2020nuscenes} validation split. We use the mean Average Precision (mAP) and the nuScenes Detection Score (NDS), where NDS aggregates multiple error types: translation (mATE), scale (mASE), orientation (mAOE), velocity (mAVE), and attribute classification (mAAE).

We use two popular yet simple image only detectors: DETR3D \cite{detr3d} and StreamPETR \cite{wang2023streampetr}, both multi-camera detectors with an EVA-02-L image encoder \cite{fang2023eva02}, six cameras, and $320\times800$ images. All temporal variants start from the same full-depth checkpoint. The four semantic exits correspond to increasing fractions of the encoder; their exact block mapping is given in \cref{app:implementation}.

We report end-to-end wall-clock latency measured during sequential evaluation and end-to-end GFLOPs under a common six-camera, $320\times800$, two-FLOPs-per-MAC convention on a single NVIDIA A100 GPU. GFLOPs are comparable across implementations that report the same boundary and convention. Absolute latency is hardware and software dependent, so speedups are normalized only to the dense baseline within each reported environment. Accuracy values are single-run measurements unless stated otherwise.

\subsection{Comparison with Prior Efficiency Methods}

\begin{table*}[t]
\label{tab:main-results}
\centering
\small
\caption{Comparison of accuracy, latency, and inference speedup on the nuScenes validation set ($320\times800$). We compare \method{} against recent state-of-the-art spatial token compression architectures. To ensure a fair evaluation across different hardware setups, we group methods by their testing environment (indicated by their dense baseline latency) and report their relative speedup$^1$ against their respective full-encoding baseline.}
\resizebox{0.9\linewidth}{!}{%
\begin{tabular}{lccrrr}
\toprule
Method & NDS & mAP & GFLOPs & Latency (ms) & Speedup\\
\midrule
\multicolumn{6}{c}{\textit{Spatial Token Compression SOTA (StreamPETR~\cite{wang2023streampetr} ViT-L Baseline)}} \\
\midrule
Dense Baseline  & 61.18 & 52.10 & 4601 & 350.6 & 1.00$\times$ \\
ToC3D-Fast \cite{zhang2024toc3d}  & 60.90 & 51.70 & 3935 & 281.0 & 1.13$\times$ \\
ToC3D-Faster \cite{zhang2024toc3d} & 60.50 & 51.30 & 3131 & 237.2 & 1.34$\times$ \\
SEPatch3D-Fast \cite{SEPatch3D} & 61.20 & 52.10 & 4199 & 250.2 & 1.27$\times$ \\
SEPatch3D-Faster \cite{SEPatch3D} & 60.30 & 51.60 & 3023 & 194.3 & 1.63$\times$ \\
SToRe3D$^2$-1/2 \cite{store3d} & 61.80 & 53.30 & - & 370.4 & 0.94$\times$ \\
SToRe3D$^2$-1/3$^2$ \cite{store3d} & 60.90 & 52.30 & - & 285.7 & 1.23$\times$ \\
SToRe3D$^2$-1/10$^2$ \cite{store3d}  & 60.70 & 52.10 & - & 192.3 & 1.82$\times$ \\
\method{} &  \underline{61.36} & \textbf{52.55} & 3735 & 182.3 & 1.92$\times$ \\

\midrule
\multicolumn{6}{c}{\textit{Temporal Depth Allocation (DETR3D~\cite{detr3d} ViT-L Baseline)}} \\
\midrule
Dense Baseline \cite{detr3d} & 60.81 & 51.2 & 5161 & 364.4 & 1.00$\times$ \\
Dense Baseline + Memory & \textbf{61.90} & \underline{52.5} & 5229 & 373.4&0.98$\times$ \\ 

\method{} &  \underline{61.5} & \textbf{52.76} & 4115 & 168.7 & 2.16$\times$ \\
\bottomrule
\end{tabular}}
\parbox{\linewidth}{\footnotesize $^1$ To account for differences in hardware, software dependencies, and warmup procedures across literature, speedups for prior works are reported relative to each method's published baseline.\\
$^2$ SToRe3D replaces the backbone with DynamicViT~\cite{rao2021dynamicvit} (slower on GPU) without releasing code or FLOPs; hence, we report wall-clock latency against the vanilla ViT-L.}

\end{table*}

\cref{tab:main-results} compares \method{} with recent efficient multi-camera
detectors and with dense baselines. Because published methods
use different detector and timing environments, speedups are
interpreted relative to the corresponding dense baseline
within each environment.

On the DETR3D configuration, \method{} reaches 61.50 NDS and 52.76 mAP at 168.7 ms, a 2.16$\times$ speedup over standard dense DETR3D, reducing GFLOPs from 5161 to 4115.
The same mechanism transfers to StreamPETR. \method{} reaches 61.36 NDS and 52.55 mAP at 182.3 ms, compared with 61.18 NDS, 52.10 mAP, and 350.6 ms for the dense baseline. This result supports
the intended claim that temporal depth allocation is
complementary to detector-side temporal modeling.

Because published methods evaluate across diverse hardware setups, we report speedups relative to the corresponding baseline within each environment.

Figure~\ref{fig:accuracy-latency} provides a complementary
view of the accuracy--efficiency trade-off. \method{} lies on
the favorable frontier of the comparison, retaining high
detection accuracy while providing substantially larger
inference acceleration than methods operating near the same
NDS range. Unlike spatial token- or patch-reduction methods,
\method{} obtains this trade-off by allocating encoder depth
temporally, making the two efficiency mechanisms
complementary rather than mutually exclusive.


\subsection{Detection Quality Beyond NDS}

\begin{table*}
\centering
\small
\caption{\textbf{Detailed nuScenes detection metrics.}
We report the standard DETR3D baseline, published SEPatch3D
operating points, and \method. Higher NDS/mAP is better;
lower true-positive errors are better. Published SEPatch3D
results use a different detector environment and are included
for context rather than as a matched controlled comparison.
Metric definitions are provided in Appendix~C.}
\label{tab:detailed_metrics}
\resizebox{0.9\linewidth}{!}{%
\begin{tabular}{l c c c c c c c}
\toprule
Method & NDS $\uparrow$ & mAP $\uparrow$ & mATE $\downarrow$ & mASE $\downarrow$ & mAOE $\downarrow$ & mAVE $\downarrow$ & mAAE $\downarrow$ \\
\midrule
DETR3D  & 60.81 & 51.19 & 0.5420 & 0.2578 & 0.2809 & 0.2440 & 0.1922 \\

SEPatch3D-Fast & 61.20 & 52.10 & 0.557 &0.253 &\textbf{0.270}& 0.236 & 0.199 \\
SEPatch3D-Faster& 60.30 & 51.60 &0.574 &0.258& 0.288 &0.236 &0.198 \\

\method{} & \textbf{61.50} & \textbf{52.76} & \textbf{0.5267} & \textbf{0.2518} & 0.2726 & \textbf{0.2351} & \textbf{0.1973} \\
\bottomrule
\end{tabular}}
\end{table*}

\cref{tab:detailed_metrics} shows that the aggregate accuracy change is not
driven by a single nuScenes error term. Relative to standard
DETR3D, \method{} improves NDS and mAP while reducing mATE,
mASE, mAOE, and mAVE; only mAAE increases slightly.
The published SEPatch3D operating points are included for
context, but because they use a different detector and
evaluation environment, we do not interpret metric-level
differences across those rows as a controlled comparison.

\subsection{Core Component Analysis}

\begin{table}
\centering
\small
\caption{Policy analysis using DETR3D on nuScenes val accuracy, computational cost, and measured latency at $320\times800$.}
\label{tab:core-ablation}
\resizebox{\linewidth}{!}{%
\begin{tabular}{lrrrrr}
\toprule
Policy & NDS & mAP & GFLOPs & Lat. (ms) & Speedup \\
\midrule
n/a & 61.90 & 52.5 & 5229 & 373.4 & 1.00$\times$ \\
 Reuse rep. & 36.58 & 21.48 & 1809.0 & 154.8 & $2.41\times$ \\
 Always \early{} & 55.75 & 45.59 & 3176.9 & 161.9 & $2.31\times$ \\
Fixed \early{} & 58.23 & 48.98 & 3220.7 & 163.5 & $2.28\times$ \\
 Fixed depth cycle & 59.30 & 50.05 & 3562.9 & 206.3 & $1.81\times$ \\
 PolicyNet & \textbf{61.50} & \textbf{52.76} & 4115 & \textbf{168.7} & \textbf{$2.21\times$} \\
\bottomrule
\end{tabular}
}
\end{table}

\cref{tab:core-ablation} isolates the roles of current evidence, temporal
context, and adaptive depth. Reusing the anchor alone is fast
but incurs a substantial accuracy loss. An \textsc{Early}
current-frame representation recovers much of this loss, and
combining the same current evidence with the full-depth anchor
provides a further 2.48 NDS and 3.39 mAP with modest latency
overhead. A fixed depth cycle provides variable compute
without content adaptation; \method{} improves both accuracy
and latency relative to this content-independent schedule,
consistent with allocating deeper computation to harder
frames. Additional fixed-depth and updater variants are
reported in the appendix.

\begin{figure*}
    \centering
    \includegraphics[width=0.85\linewidth]{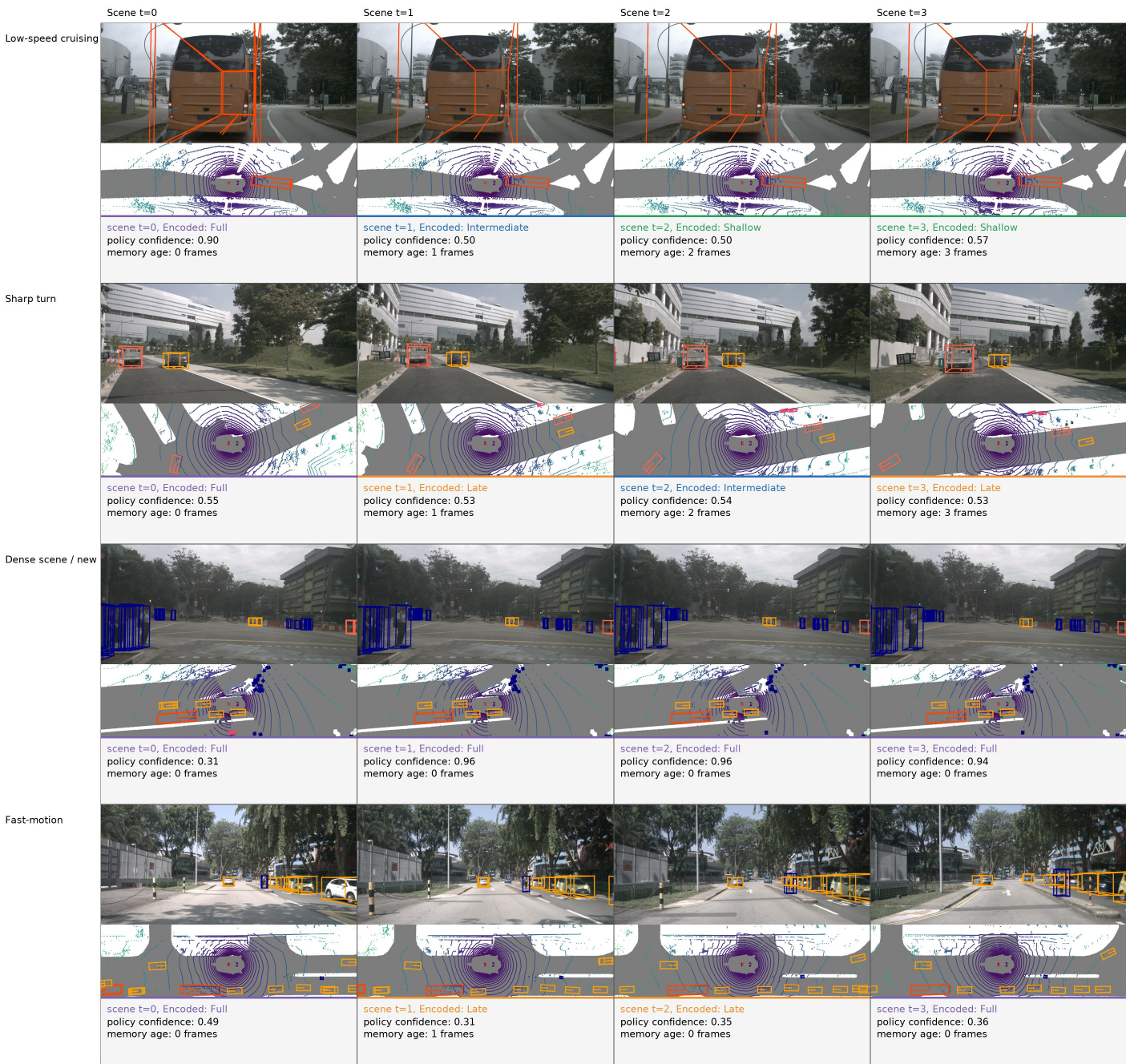}
\caption{\textbf{Learned behavioral regimes of the pre-encoding controller.} Each row displays a representative driving sequence from one of the four behavioral clusters. Each timestamp shows the front-camera view, BEV prediction, selected encoder depth, policy confidence, and anchor age.}
    \label{fig:qualitative_routing}
\end{figure*}

\subsection{Predict the Change Before Encoding}

We analyze adjacent validation frames with scene-grouped cross-validation. Online metadata predicts full-feature similarity, and a current \early{} feature probe improves prediction further (\cref{tab:signal-regression}). The probe is diagnostic rather than part of the headline policy because evaluating it before routing changes the latency accounting.
Additional diagnostics in the appendix figures show the target
distribution, the contribution of individual online signals,
and the relationship between early and final representation
change.

\begin{table}
\centering
\small
\caption{Scene-grouped regression of adjacent full-feature cosine similarity. Values are mean $\pm$ standard deviation across five folds. Offline metadata uses annotations and is an oracle.}
\label{tab:signal-regression}
\resizebox{\columnwidth}{!}{%
\begin{tabular}{lrr}
\toprule
Policy features & $R^2$ & MAE \\
\midrule
Online metadata & $0.706\pm0.043$ & $0.0391\pm0.0020$ \\
Online + current \early{} probe & $0.902\pm0.021$ & $0.0216\pm0.0010$ \\
Offline metadata & $0.853\pm0.037$ & $0.0268\pm0.0014$ \\
Offline + current \early{} probe & $0.952\pm0.010$ & $0.0149\pm0.0006$ \\
\bottomrule
\end{tabular}}
\end{table}




\begin{table}
\centering
\small
\caption{Hardware footprint and resource profiling on nuScenes val. We report end-to-end per-frame latency, peak CUDA allocated/reserved memory (MB), and computational cost under the
same six-camera, $320\times800$ eval protocol.}
\label{tab:memory-latency-profile}
\resizebox{\linewidth}{!}{%
\begin{tabular}{lcccc}
\toprule
Configuration & Lat. (ms) $\downarrow$ & Peak Alloc. $\downarrow$ & Peak Res. $\downarrow$ & GFLOPs $\downarrow$ \\
\midrule
DETR3D  & 373.40 & \textbf{1781} & \textbf{2850} & 5229 \\ 
\method{}    & \textbf{168.71} & 1918 & 2716 & \textbf{4115} \\
\bottomrule
\end{tabular}}
\end{table}

\subsection{Memory Overhead of Feature Caching}
\label{sec:memory_overhead}

As summarized in Tab.~\ref{tab:memory-latency-profile}, the persistent appearance anchor
introduces only a modest allocated-memory overhead. Relative to the matched full-depth configuration, peak allocated memory increases from 1781.36 MB to 1917.95 MB
(+7.67\%), while end-to-end latency and computation decreases segnificantly. Peak reserved memory decreases from 2850 to
2716 MB; because reserved memory is controlled by the CUDA
caching allocator, we use peak allocated memory as the more
meaningful footprint measure.






\subsection{What Does the Controller Learn: Behavioral Clustering \& Routing}
\label{sec:qualitative_routing}

To investigate what the pre-encoding controller learns beyond aggregate accuracy--latency trade-offs, we analyze the structure of the policy's output space across $700$ training clips. We project the predicted policy logits into a 2D embedding via t-SNE and partition them into four clusters (App.\ Fig.~\ref{fig:policy_tsne}). A manual audit reveals that these four clusters naturally align with distinct physical driving regimes: (1) \emph{low-speed cruising}, (2) \emph{sharp turns}, (3) \emph{dense scenes / new objects}, and (4) \emph{fast motion} (see App.\.~\ref{sec:appendix_policy_audit} for audit details).

Fig.~\ref{fig:qualitative_routing} illustrates a representative sequence from each discovered behavioral cluster. \textsc{Full} encoding is enforced at scene starts by the validity safeguards, after which stable \emph{low-speed cruising} permits progressively shallower exits (\textsc{Intermediate} $\rightarrow$ \textsc{Early}) as the full-depth anchor ages. \emph{Sharp turns} induce deeper intermediate exits (\textsc{Intermediate}/\textsc{Late}) to compensate for rapid viewpoint rotation, while \emph{dense scenes} with heavy occlusion and entering agents remain at \textsc{Full} depth. In the \emph{fast-motion} regime, the controller likewise favors deeper computation (\textsc{Late}/\textsc{Full}) and exhibits lower confidence on intermediate decisions. Rather than relying on hand-crafted rules, the controller autonomously organizes temporal kinematics and scene summaries into meaningful semantic routing regimes: allocating deeper computation when temporal reuse is unreliable and retreating to shallow prefixes when high-level appearance remains informative.

\section{Limitations}

The main evaluation uses one dataset, one image resolution, one detector family, and one large transformer backbone. The adapter interface has been implemented for a second detector family, but a completed transfer experiment is still missing. Most accuracy rows are single runs, and the latest operating point lacks a complete action-weighted operation-count record and repeated latency statistics.

The full-representation similarity target is not task aware. Small objects can be important even when their effect on the global feature average is weak. The teacher-forced policy also sees cleaner preceding object summaries during training than at inference. Feature noise reduces but does not remove this shift. Finally, the cached appearance feature is defined on camera image grids and is not explicitly aligned under motion. The method relies on current partial evidence and periodic full refreshes instead of geometric feature warping.

\section{Conclusion}

\method{} reduces repeated image-backbone computation by
pairing a read-only full-depth appearance anchor with
policy-selected partial features from the current frame.
Current calibration and the motion-conditioned positional
adapter keep geometric conditioning frame-specific, while the
controller allocates encoder depth before image processing.
Across DETR3D and StreamPETR configurations on nuScenes, the
method preserves strong detection accuracy while
substantially reducing latency, achieving a 2.16$\times$
speedup over DETR3D and a 1.92$\times$
speedup over StreamPETR. Controlled comparisons
show distinct contributions from current evidence, full-depth
anchoring, lightweight correction, and adaptive routing.
The principal limitations are the absence of explicit spatial
warping, hardware-dependent absolute latency, and evaluation
on a single dataset.

\clearpage

{\small
\bibliographystyle{plainnat}
\bibliography{main}
}
\clearpage
\appendix
\begin{center} {\Large\bfseries Supplementary Material - EvoMem3D} \end{center} 

\section{Implementation Details}
\label{app:implementation}

\paragraph{Semantic exit mapping.}
The main paper uses relative encoder-depth names so that the method is not tied to one backbone. In the reported EVA-02-L implementation, \early{}, \intermediate{}, \lateexit{}, and \fullexit{} correspond to blocks 8, 12, 16, and 24, respectively. Partial features are projected through separate four-scale feature pyramids to tensors of shape $B\times V\times256\times H_s\times W_s$.

\paragraph{Updater architecture.}
At each scale, the gated residual updater concatenates the cached full feature and projected partial feature, then applies separate $1\times1$ convolutions for the sigmoid gate and residual. A learned scalar residual scale is initialized to zero. The default implementation shares this updater across partial exits and uses an exit-specific projector.

\paragraph{Policy network and labels.}
The reported policy is a two-layer MLP with a 32-dimensional hidden layer and four output classes. The deployable input vector contains timestamp and ego-motion history plus summaries of the preceding detector output. The strongest configuration uses 42 scalar inputs. Full-feature similarity thresholds map training samples to the four semantic exits; the base thresholds are:
$$
y_t=
\begin{cases}
\textsc{Early}, & s_t\ge 0.90,\\
\textsc{Intermediate}, & 0.76\le s_t<0.90,\\
\textsc{Late}, & 0.675\le s_t<0.76,\\
\textsc{Full}, & s_t<0.675.
\end{cases}
$$
with class weights $(1,1,1,3)$. These values are implementation choices of the reported configuration, not part of the general method.

\paragraph{Training.}
Temporal runs initialize from the trained DETR3D EVA-02 checkpoint and freeze pretrained detector parameters. The inherited detector optimizer is AdamW with learning rate $4\times10^{-4}$, backbone multiplier $0.1$, weight decay $0.01$, cosine annealing, 500 warm-up iterations, mixed precision, and gradient clipping at norm 35. Training runs for 42,192 iterations. Policy similarity supervision is active for the first 10\% of training. The strongest row uses Gaussian noise with standard deviation $0.10$ in normalized policy-feature units.

\paragraph{Teacher forcing and inference.}
For object-summary features, training uses the preceding ground-truth objects and inference uses preceding detector predictions. Current annotations are never used by the deployable policy. The full feature path is used without gradient to construct policy labels during policy warm-up and, when enabled, as a distillation teacher. Assertions prohibit a teacher call on partial frames during inference.

\paragraph{Validity safeguards.}
At scene starts and sequence discontinuities, the next frame uses the full path. In the reported configuration, the anchor age is limited to four frames and to two frames when ego speed exceeds $2.0$ m/s or the preceding predicted maximum object speed exceeds $5.0$ m/s. These are inference safeguards against stale context rather than architectural requirements.

\section{Extended Controlled Results}

\begin{table*}[t]
\centering
\small
\caption{\textbf{Controlled depth-routing comparison.}
All methods use the same detector checkpoint and refresh constraints.
``Partial only'' uses the current partial representation without the
cached anchor, whereas ``Anchor update'' combines it with the latest
full-depth anchor.}
\label{tab:fixed-extended}
\resizebox{\textwidth}{!}{%
\begin{tabular}{llrrrrrr}
\toprule
Routing & Non-full-frame computation & NDS & mAP & GFLOPs & Latency (ms) & FPS & Speedup \\
\midrule
Full depth & - & \textbf{61.90} & \underline{52.50} & 5229 & 373.4 & 2.68 & $1.00\times$ \\
Naive reuse & Anchor only & 36.58 & 21.48 & 1809.0 & 104.8 & 9.54 & $3.55\times$ \\
Partial only & \early{} & 55.75 & 45.59 & 3176.9 & 121.9 & 8.2 & $3.06\times$ \\
Partial only & \intermediate{} & 59.46 & 50.55 & 3715.8 & 197.9 & 5.05 & $1.89\times$ \\
Partial only & \lateexit{} & 60.81 & 51.96 & 4254.6 & 236.1 & 4.24 & $1.58\times$ \\
Memory update & Fixed \early{} & 58.23 & 48.98 & 3220.7 & 163.5 & 6.18 & $2.28\times$ \\
Memory update & Fixed depth cycle & 59.30 & 50.05 & 3562.9 & 206.3 & 4.85 & $1.81\times$ \\
\method{} & Adaptive semantic exit & \underline{61.50} & \textbf{52.76} & 4115 & 168.7 & 5.93 & $2.16\times$ \\
\bottomrule
\end{tabular}}
\end{table*}

\Cref{tab:fixed-extended} expands the fixed-depth and schedule comparison from the main paper, and \cref{tab:controller_ablation} reports all completed learned-controller experiments; the updater ablation appears in (\cref{tab:updater-ablation}).

\begin{table*}[t]
\centering
\caption{\textbf{Controller ablations.}
(a) isolates controller information and inference-time constraints.
Ground-truth rows are oracle diagnostics and are not deployable.
(b) evaluates the training recipe using deployable previous-frame
predictions at inference.}
\label{tab:controller_ablation}
\small

\begin{tabular}{l l c c c c }
\toprule
\multicolumn{6}{c}{\textbf{(a) Controller signal and constraints}}\\
\midrule
Variant & Test signal & Age bound & 
NDS & mAP  & Lat. (ms)\\
\midrule
Online & Metadata & \xmark & 
55.21 & 44.86  & 226.91\\
Online + age bound & Metadata & \cmark & 
57.94 & 48.57  & 212.28\\
Current-GT oracle & Current GT &  -- & 
57.17 & 47.24 & --\\
Previous-GT oracle & Previous GT  & -- &
57.38 & 47.48 & --\\
Prev.-output summary & Previous prediction & \cmark & 
60.05 & 50.61 & 193.35\\
\bottomrule
\end{tabular}


 \begin{tabular}{l c c c c c c}
  \toprule
  \multicolumn{7}{c}{\textbf{(b) Deployable training recipe}}\\
  \midrule
  Variant & Feature noise & Motion guard &
  NDS & mAP & \%FULL & Lat. (ms)\\
  \midrule
  Base controller & \xmark & \xmark &
  60.05 & 50.61 & 44.87 & 193.35\\
  + class weighting & \xmark & \xmark &
  60.43 & 51.36 & 48.10 & 189.58\\
  + feature noise & \cmark & \xmark &
  60.57 & 51.39 & 48.13 & 190.78\\
  + motion guard & \cmark & \cmark &
  \textbf{61.49} & \textbf{52.77} & 61.99 & \textbf{179.83}\\
  + Motion PE  Correction & \cmark & \cmark &
  \textbf{61.50} & 52.76 & 59.51 & \textbf{168.71}\\
  + distillation & \cmark & \cmark &
  61.39 & 52.53 & 61.89 & 180.91\\
  \bottomrule
  \end{tabular}
\end{table*}

\begin{figure*}
    \centering
    \begin{minipage}[t]{0.49\textwidth}
        \centering
        \includegraphics[width=\linewidth]{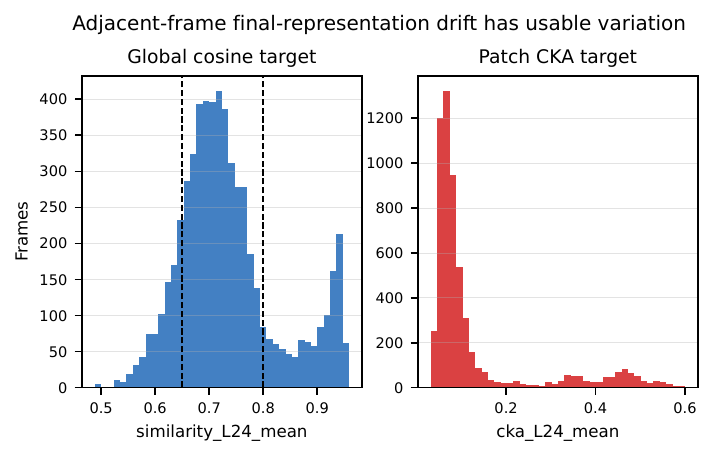}
    \end{minipage}\hfill
    \begin{minipage}[t]{0.49\textwidth}
        \centering
        \includegraphics[width=\linewidth]{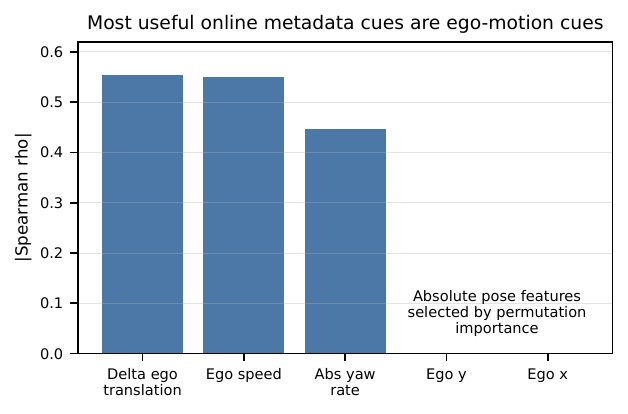}
    \end{minipage}
    \caption{\textbf{Additional policy-signal diagnostics.} \textbf{Left:} distributions of the adjacent-frame similarity targets (global cosine and patch CKA) used to derive depth labels. \textbf{Right:} the online metadata cues most correlated with representation change; ego-motion magnitudes dominate. These associations are predictive, not causal.}
    \label{fig:policy-diagnostics}
\end{figure*}

\section{Detection Metrics and Per-Class Results}
\label{app:per_class}

\paragraph{Detailed nuScenes metrics.}
nuScenes reports mAP over the ten detection classes and standard
center-distance thresholds, together with five mean true-positive
errors: mATE measures center translation error in meters; mASE
measures scale error as $1-\mathrm{IoU}$ after aligning translation
and orientation; mAOE measures yaw-orientation error in radians; mAVE
measures velocity error in $\mathrm{m/s}$; and mAAE measures attribute
error as one minus attribute accuracy. NDS combines mAP with
normalized versions of these five errors, with mAP contributing half
of the overall score. Thus, higher mAP and NDS are better, whereas
lower mATE, mASE, mAOE, mAVE, and mAAE are better.

Table~\ref{tab:per-class-detailed} reports the per-class metrics for \method. The mean
of the ten AP values is 52.76, matching the aggregate mAP in the main text.
Because the table contains a single run, we report these values descriptively
and do not interpret small category-level differences as statistically
significant.


\begin{table}
\centering
\small
\caption{Detailed nuScenes validation results by class. AP is averaged over the standard center-distance thresholds and reported as a percentage; true-positive errors are reported in their native fractional units. A dash denotes a metric that nuScenes does not define: AOE, AVE, and AAE for traffic cones, and AVE and AAE for barriers.}
\label{tab:per-class-detailed}
\resizebox{\linewidth}{!}{%
\begin{tabular}{lrrrrrr}
\toprule
Class & AP & ATE & ASE & AOE & AVE & AAE \\
\midrule
Car                  & 69.30 & 0.367 & 0.145 & 0.056 & 0.225 & 0.189 \\
Truck                & 47.73 & 0.575 & 0.191 & 0.067 & 0.195 & 0.205 \\
Construction vehicle & 21.08 & 0.846 & 0.414 & 0.873 & 0.133 & 0.361 \\
Bus                  & 55.58 & 0.623 & 0.189 & 0.049 & 0.418 & 0.281 \\
Trailer              & 28.85 & 0.863 & 0.229 & 0.341 & 0.165 & 0.164 \\
Barrier              & 64.54 & 0.323 & 0.279 & 0.141 & --    & --    \\
Motorcycle           & 55.39 & 0.477 & 0.244 & 0.255 & 0.343 & 0.228 \\
Bicycle              & 56.21 & 0.400 & 0.245 & 0.338 & 0.135 & 0.009 \\
Pedestrian           & 59.05 & 0.517 & 0.284 & 0.353 & 0.265 & 0.141 \\
Traffic cone         & 69.86 & 0.288 & 0.303 & --    & --    & --    \\
\bottomrule
\end{tabular}}
\end{table}


The aggregate mAP difference is spread across categories rather than driven by a single class. Relative to the DETR3D baseline, the largest gains are for construction vehicles ($+1.96$ AP), trailers ($+1.57$), motorcycles ($+1.48$), and bicycles ($+1.23$), while trucks decrease slightly ($-0.22$). Per-class differences require repeated runs before they can support category-specific claims.

\section{Representation Drift, Depth Demand, and Reuse Risk}
\label{app:drift-demand}

\begin{figure}[t]
    \centering
    \includegraphics[width=\linewidth]{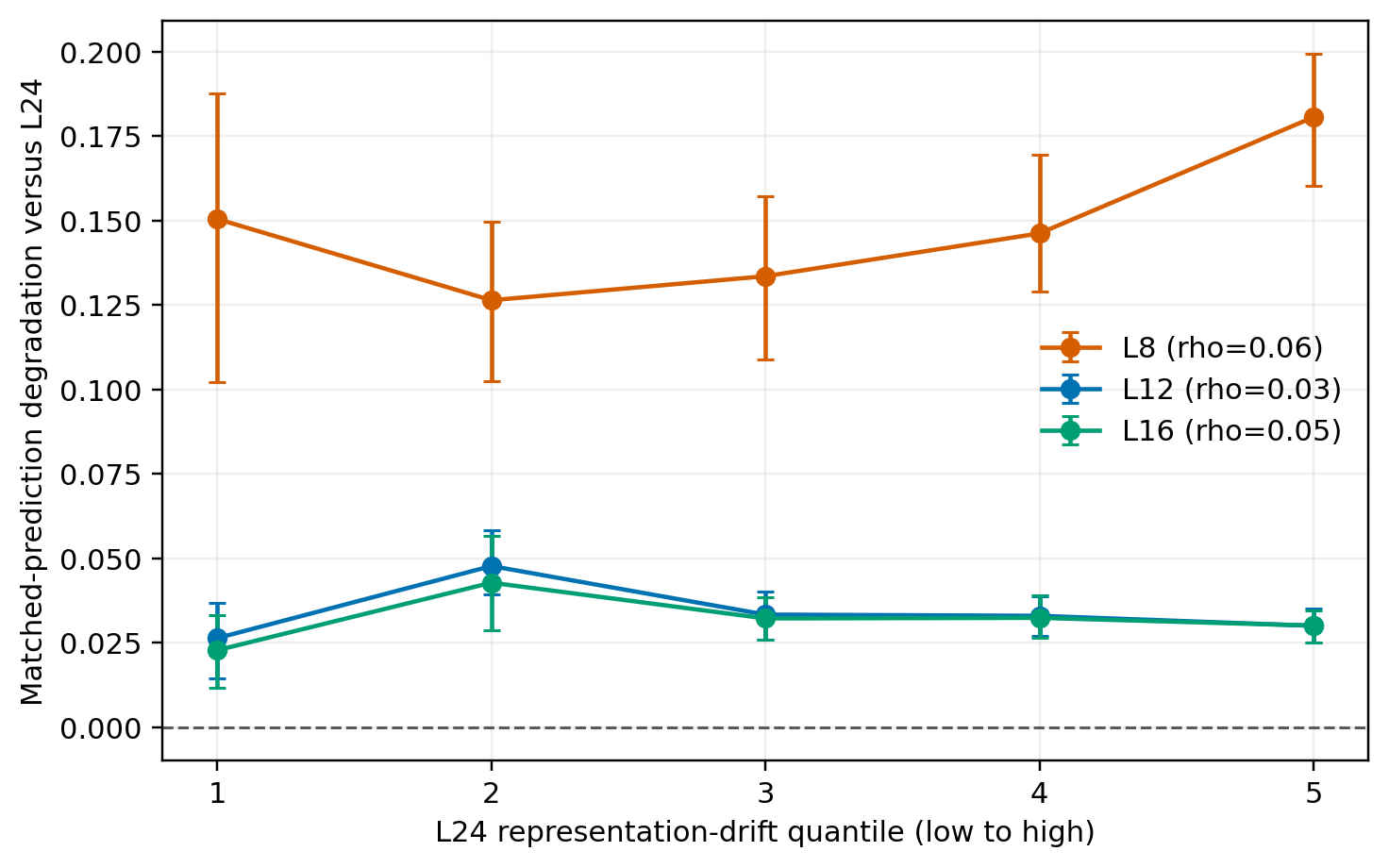}
    \caption{\textbf{Full-feature drift versus matched-depth degradation.}
    Five equal-count drift groups (low to high) show mean L8, L12, and
    L16 error minus matched L24 error. Bars are scene-bootstrap 95\%
    intervals; $\rho$ denotes frame-level Spearman correlation. Depth
    matters more than the weak drift trend within a depth.}
    \label{fig:drift-vs-depth-quantiles}
\end{figure}

\begin{figure*}[t]
    \centering
    \includegraphics[width=\textwidth]{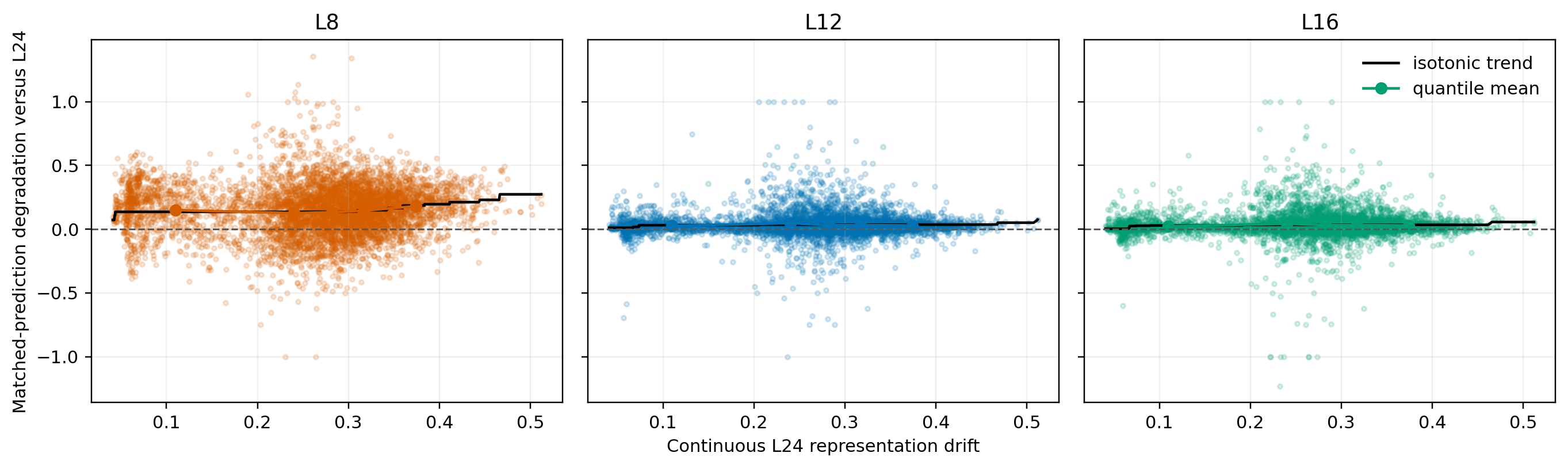}
    \caption{\textbf{Frame-level matched-depth counterfactuals.}
    Each point is one validation frame evaluated from a shared pre-frame
    state. Colored markers show drift-quintile means and black lines show
    isotonic fits. The wide scatter and small correlations caution against
    treating full-feature drift as a per-frame detection-regret oracle.}
    \label{fig:drift-vs-depth-scatter}
\end{figure*}

\paragraph{Matched-depth counterfactuals.}
We compare L8, L12, and L16 actions with L24 from the same pre-frame
temporal state on 5,869 nuScenes validation frames in 150 scenes. For
each action $a$, we define degradation as
\begin{equation}
    \Delta_t^a=\mathcal{E}(\widehat{Y}_t^a,Y_t)
    -\mathcal{E}(\widehat{Y}_t^{\mathrm{L24}},Y_t).
\end{equation}
Here $\mathcal{E}$ is a class-aware matched-prediction error combining
confidence, center, size, yaw, velocity, and unmatched-box penalties;
it is a diagnostic surrogate, not nuScenes mAP or NDS. The drift
measure is adjacent-frame cosine change in the full L24 representation.
Figure~\ref{fig:drift-vs-depth-quantiles} shows a larger average penalty
for L8 than for L12 or L16. Within each depth, however, drift is only
weakly associated with degradation: Spearman $\rho$ is 0.062, 0.032,
and 0.048 for L8, L12, and L16, respectively, and each scene-bootstrap
95\% interval includes zero. The L8 quintile means are also not
monotonic. Figure~\ref{fig:drift-vs-depth-scatter} exposes the broad
frame-level variation behind those averages. These observations limit
the use of global drift as a proxy for the task-level cost of a shallow
exit; they do not establish a frame-level safe-reuse threshold.

\begin{figure*}[t]
    \centering
    \includegraphics[width=\textwidth]{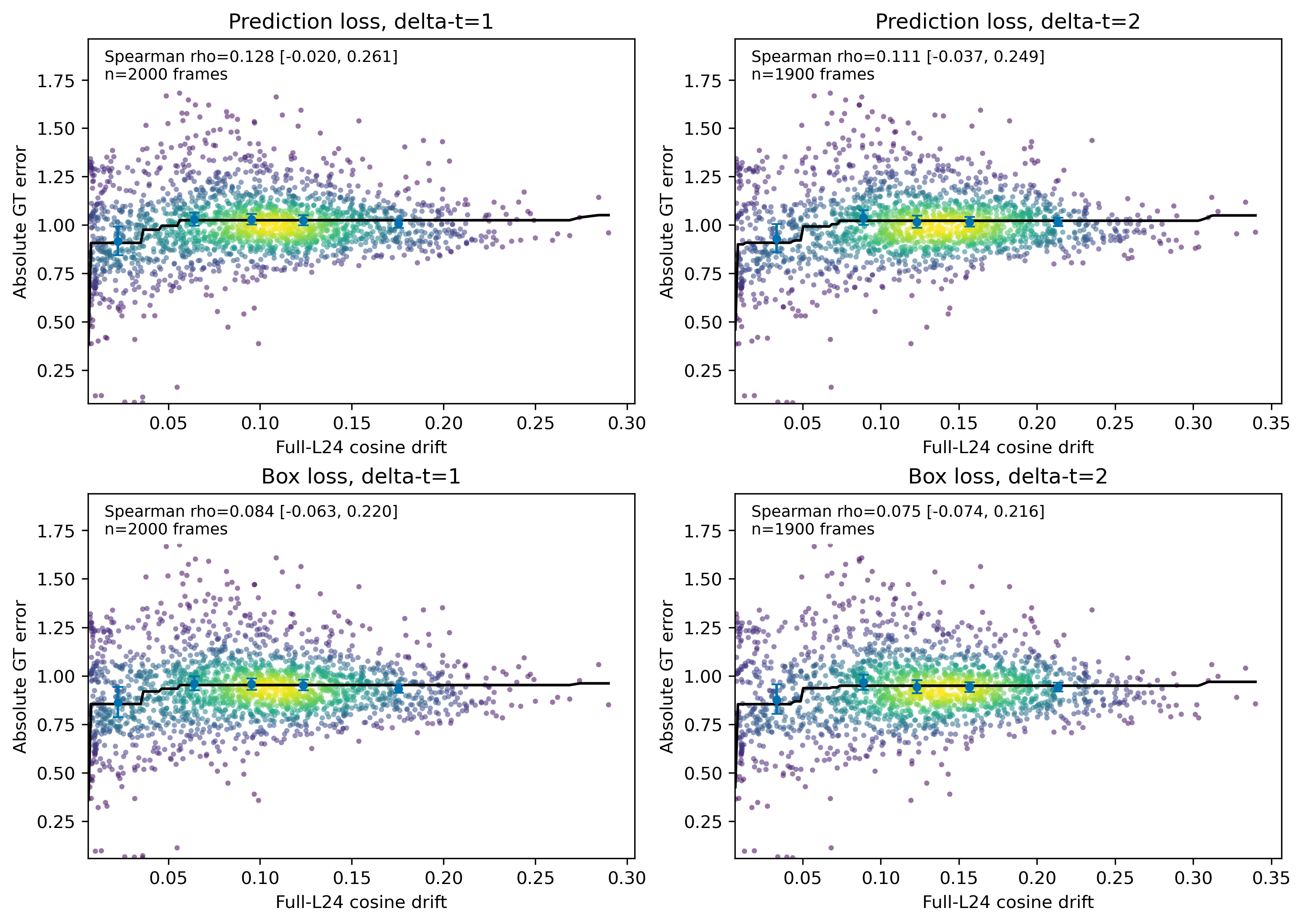}
    \caption{\textbf{Full-L24 drift and absolute error under naive reuse.}
    Top: matched-prediction error; bottom: box-only error. Columns use
    one- and two-frame drift offsets (2,000 and 1,900 frames). Blue
    markers are drift-quintile means with scene-bootstrap 95\% intervals;
    black lines are isotonic fits. Reported $\rho$ intervals all include
    zero. Absolute error is not reuse-induced degradation.}
    \label{fig:drift-vs-reuse-loss}
\end{figure*}

\paragraph{Naive-reuse risk.}
We separately examine the deployed naive-reuse $K=2$ detector on 2,000
non-keyframes from 100 scenes. Figure~\ref{fig:drift-vs-reuse-loss}
compares full-L24 cosine drift over one and two frame offsets with
\emph{absolute} ground-truth matched-prediction and box errors. The
prediction-error correlations are $\rho=0.128$ and $0.111$ for the two
offsets; the box-error correlations are $0.084$ and $0.075$. All four
scene-bootstrap 95\% intervals include zero. Because these outcomes
include intrinsic frame difficulty and have no matched full-encoding
counterfactual, this analysis cannot estimate error caused specifically
by reuse. It provides no reliable drift-only rule for deciding when
naive reuse is safe.

\section{Update-Operator Variants}
\label{app:updaters}

Table~\ref{tab:updater-ablation} compares the gated residual updater with
attention-based alternatives under the same \textsc{Early} partial features
and fixed full-refresh schedule. The gated residual provides the best accuracy
among the evaluated operators while matching the fastest measured throughput.

\begin{table}
\centering
\small 
\caption{Updater ablation with L8 partial features and a fixed full-encoding schedule. A lightweight gated residual yields the best accuracy without the latency overhead of cross-attention.}
\label{tab:updater-ablation}
\begin{tabular}{@{}lccc@{}}
\toprule
Non-full-frame operation & NDS & mAP & FPS \\
\midrule
Partial only & 55.75 & 45.59 & 6.12 \\
Gated residual & \textbf{58.23} & \textbf{48.98} & \textbf{6.12} \\
Cross-attention & 54.33 & 43.58 & 5.95 \\
Gated cross-attention & 57.61 & 48.30 & 5.91 \\
Per-camera attention & 55.04 & 44.30 & 5.99 \\
Inverted attention & 52.91 & 42.49 & 5.90 \\
Gated residual + distillation & 58.08 & 48.69 & 6.12 \\
\bottomrule
\end{tabular}
\end{table}

\begin{figure*}
    \centering
    \begin{minipage}[t]{0.49\textwidth}
        \centering
        \includegraphics[width=0.8\linewidth]{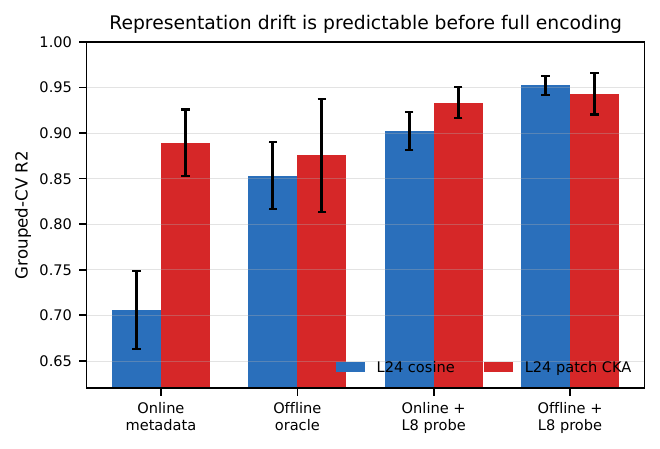}
    \end{minipage}\hfill
    \begin{minipage}[t]{0.49\textwidth}
        \centering
        \includegraphics[width=0.8\linewidth]{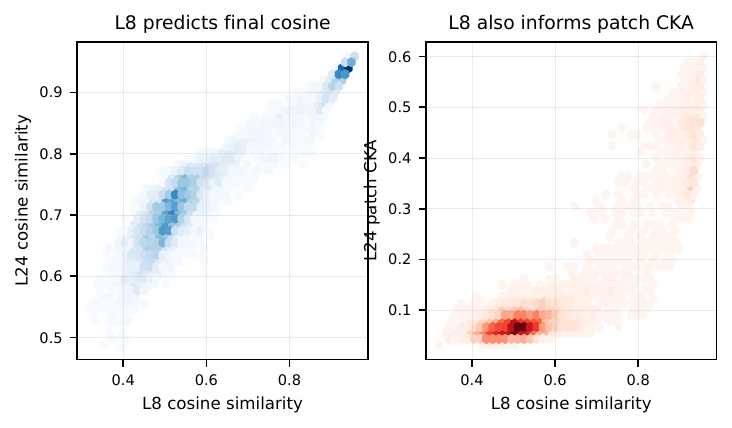}
    \end{minipage}
    \caption{Evidence for controller inputs. \textbf{Left:} scene-grouped prediction of full-representation change. \textbf{Right:} the current early representation is informative about final-feature change. A cascade that uses this probe must include its computation in end-to-end latency.}
    \label{fig:policy-signals}
\end{figure*}

\section{Detailed Inference Profiling}
\label{app:latency}

\begin{figure}
    \centering
    \includegraphics[width=\linewidth]{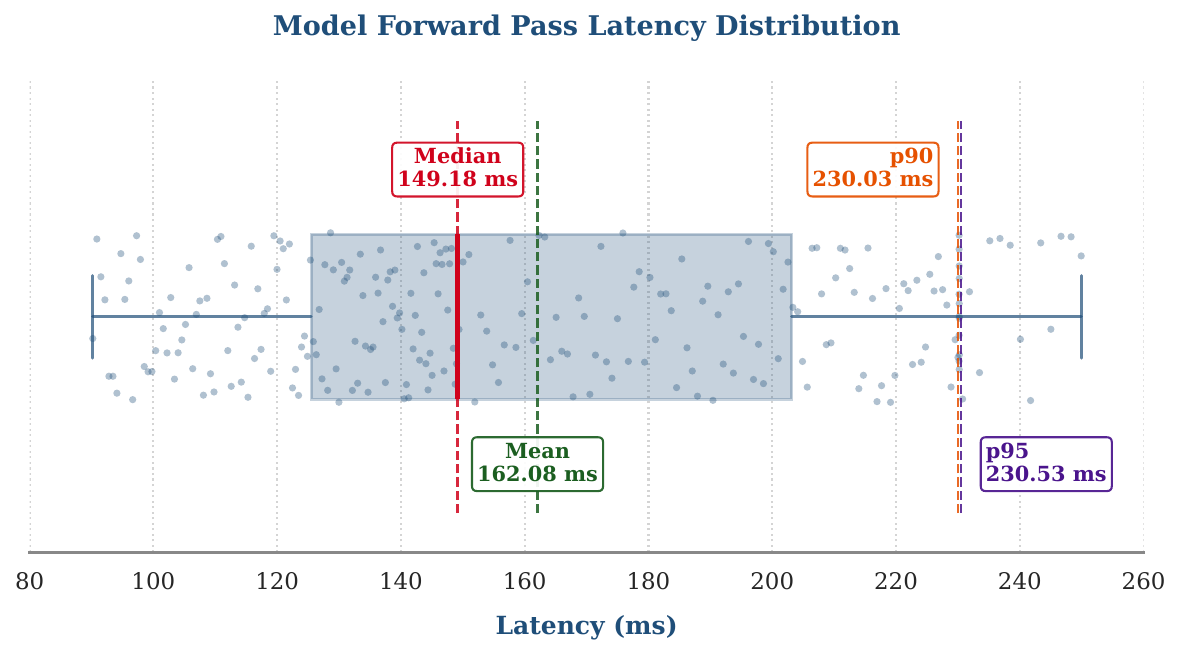}
    \caption{Forward pass latency distribution on the nuScenes validation split using the StreamPETR model. The adaptive routing policy maintains a highly efficient median latency of $149.18$ ms, while effectively bounding tail latencies (${\sim}230$ ms) during full-depth feature refreshes. Peak CUDA memory footprint during these operations requires $3.37$ GB allocated and $16.69$ GB reserved.}
    \label{fig:latency_distribution}
\end{figure}

We report two timing scopes. \emph{End-to-end sequential latency} includes
fixed exit strategies and is the quantity used in
Tables~\ref{tab:fixed-extended} and~\ref{tab:controller_ablation}. \emph{Forward latency}
measures only the synchronized StreamPETR model forward pass and using our approach in Fig.~\ref{fig:latency_distribution}; it is therefore lower and should not be
compared directly with the end-to-end values.

All measurements use batch size one on A100 GPU, with FP16,
Each configuration is evaluated in a fresh process. We discard the first 2 warm-up frames, call \texttt{torch.cuda.synchronize()} before and after each timed region, and
measure 1,000 ordered nuScenes validation frames.





\section{Policy Logit Clustering and Behavioral Audit}
\label{sec:appendix_policy_audit}

\begin{figure}
    \centering
    \includegraphics[width=\linewidth]{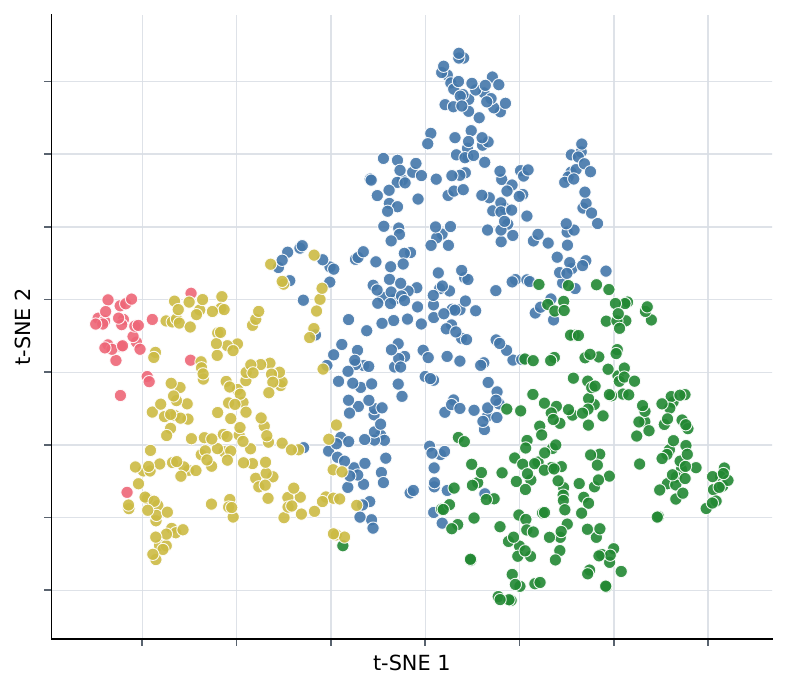}
\caption{\textbf{t-SNE visualization and clustering of learned policy logits.} We project the pre-encoding controller's output logits across $700$ training clips into two dimensions using t-SNE and partition them into four clusters (color-coded). A manual audit of $10\%$ of the data reveals that the policy organizes driving dynamics into four interpretable regimes without explicit behavioral labels: \emph{low-speed cruising}, \emph{sharp turns}, \emph{dense scenes}, and \emph{fast motion} (qualitative examples shown in Fig.~\ref{fig:qualitative_routing}).}
\label{fig:policy_tsne}
\end{figure}

To verify what underlying scene dynamics the pre-encoding controller captures from the deployable input vector $q_t$ (timestamp, ego-motion history, and preceding detection summaries), we perform an unsupervised clustering analysis of its decision space. Specifically, we extract the pre-softmax policy logits $\pi_\theta(a_t \mid q_t) \in \mathbb{R}^4$ across $700$ training clips from nuScenes and project them into a 2D embedding space via t-SNE, applying $k$-means clustering ($k=4$), as shown in Fig.~\ref{fig:policy_tsne}.

Because the policy is supervised solely by full-representation cosine similarity $s_t$ rather than explicit driving-scenario labels, we conducted a manual blind audit on a stratified $10\%$ sample  drawn across the four clusters. By inspecting the multi-camera video streams, ego-vehicle kinematics, and surrounding object density, we found that the four clusters consistently map to distinct physical driving behaviors:
\begin{itemize}
    \item \textbf{Low-Speed Cruising:} Steady straight-line driving or gentle coasting in low-to-moderate traffic, where visual drift is minimal and the policy predominantly routes frames to \textsc{Early} and \textsc{Intermediate} exits.
    \item \textbf{Sharp Turns:} High yaw-rate maneuvers at intersections or curves, where rapid perspective shift across cameras triggers \textsc{Intermediate} and \textsc{Late} exits to refresh appearance features.
    \item \textbf{Dense Scenes / New Objects:} Crowded urban intersections, heavy occlusions, or rapidly entering nearby agents, where preceding object-summary cues and high representation drift cause the controller to sustain \textsc{Full} encoding.
    \item \textbf{Fast Motion:} High ego translational velocity where scene geometry evolves rapidly, prompting deeper exits (\textsc{Late}/\textsc{Full}) alongside tighter validity safeguards.
\end{itemize}
This audit confirms that supervising the lightweight MLP controller with representation-change thresholds enables it to implicitly discover and separate semantically meaningful driving regimes at inference time.

\end{document}